\documentclass[journal]{IEEEtran}
\usepackage{amsmath,amsfonts}
\usepackage{bbm}
\usepackage{url}
\usepackage{arydshln}
\usepackage{booktabs}
\usepackage[flushleft]{threeparttable}
\usepackage{color}
\usepackage{listings}
\usepackage[absolute]{textpos}
\usepackage{hyperref}

\newcommand{\tcb}[1]{{\textcolor{blue}{#1}}}

\newcommand{\SubItem}[1]{
    {\setlength\itemindent{15pt} \item[-] #1}
}

\ifCLASSINFOpdf
  \usepackage[pdftex]{graphicx}
\else
  \usepackage[dvips]{graphicx}
\fi
\begin{document}
%

\title{
A Physics-Informed Deep Learning Paradigm for Traffic State 
and Fundamental Diagram Estimation}

%
%
%

\author{Rongye~Shi,~\IEEEmembership{Member,~IEEE,}
        Zhaobin Mo, Kuang Huang, Xuan Di,~\IEEEmembership{Member,~IEEE,}
        and~Qiang~Du
\thanks{Manuscript received March 19, 2020; revised November 9, 2020, January 23, 2021, June 5, 2021 and August 17, 2021; accepted
August 17, 2021.
\textit{(Corresponding author: Xuan Di.)}}
\thanks{Rongye Shi, Zhaobin Mo, and Xuan Di are with the Department of Civil Engineering and Engineering Mechanics, Columbia University, New York, NY, 10027 USA (e-mail: rongyes@alumni.cmu.edu; zm2302@columbia.edu; sharon.di@columbia.edu).} 
\thanks{Kuang Huang and Qiang Du are with the Department of Applied Physics and Applied Mathematics, Columbia University, New York, NY, 10027 USA (e-mail: kh2862@columbia.edu; qd2125@columbia.edu).} 
\thanks{Xuan Di and Qiang Du are also with the Data Science Institute, Columbia University, New York, NY, 10027 USA.}
}

%
%
\begin{textblock}{17}(2,0.1)
\tcb{This preprint is originally revised and extended from its previous version (\href{https://arxiv.org/abs/2101.06580v1}{https://arxiv.org/abs/2101.06580v1}), so there may exist some overlaps. However, due to significant differences, both are kept separately on ArXiv. It is recommended to read and make use of these two preprints complementarily. }
\tcb{Please cite this paper as: \\
Shi, Rongye, et al.``A Physics-Informed Deep Learning Paradigm for Traffic State and Fundamental Diagram Estimation." IEEE Transactions on Intelligent Transportation Systems (2021). DOI: 10.1109/TITS.2021.3106259}
\end{textblock}

%



\maketitle

\begin{abstract}

Traffic state estimation (TSE) bifurcates into two categories, model-driven and data-driven (e.g., machine learning, ML), while each suffers from either deficient physics or small data. To mitigate these limitations, recent studies introduced a hybrid paradigm, physics-informed deep learning (PIDL), which contains both model-driven and data-driven components. This paper contributes an improved version, called physics-informed deep learning with a fundamental diagram learner (PIDL+FDL), which integrates ML terms into the model-driven component to learn a functional form of a fundamental diagram (FD), i.e., a mapping from traffic density to flow or velocity. 
The proposed PIDL+FDL has the advantages of performing the TSE learning, model parameter identification, and FD estimation simultaneously. 
We demonstrate the use of PIDL+FDL to solve popular first-order and second-order traffic flow models and reconstruct the FD relation as well as model parameters that are outside the FD terms. 
We then evaluate the PIDL+FDL-based TSE using the Next Generation SIMulation (NGSIM) dataset. The experimental results show the superiority of the PIDL+FDL in terms of improved estimation accuracy and data efficiency over advanced baseline TSE methods, and additionally, the capacity to properly learn the unknown underlying FD relation.

\end{abstract}


%
\IEEEpeerreviewmaketitle

\section{Introduction}
\label{sec-1-r3}
\IEEEPARstart{T}{traffic} state estimation (TSE) refers to the data mining problem of reconstructing traffic state variables, including but not limited to flow, 
density, 
and speed, 
on road segments using partially observed data from traffic sensors~\cite{Seo-17}. 
TSE approaches can be briefly divided into two main categories: model-driven and data-driven ~\cite{di2021survey}. 
A model-driven approach is based on a priori knowledge of traffic dynamics, usually described by a physical model, e.g., the Lighthill-Whitham-Richards (LWR) model~\cite{Lighthill-1955, Richards-1956} and Aw-Rascle-Zhang (ARZ) model~\cite{Aw-Rascle-2000,Zhang-trb-2002}, to estimate the traffic state using partial observation. It assumes the model to be representative of the real-world traffic dynamics such that the unobserved values can be properly added using the model with small data. The disadvantage is that existing models, which are provided by different modelers, may only capture limited dynamics of the real-world traffic, resulting in low-quality estimation in the case of inappropriately-chosen models and poor model calibrations. Paradoxically, sometimes, verifying or calibrating a model requires a large amount of observed data, undermining the data efficiency of model-driven approaches. 

A data-driven approach is to infer traffic states based on the dependence learned from historical data using statistical or machine learning (ML) methods. Approaches of this type do not use any explicit traffic models or other theoretical assumptions, and can be treated as a ``black box" with no interpretable and deductive insights. The disadvantage is that in order to maintain a good generalizable inference to long-term unobserved values, massive and representative historical data are a prerequisite, leading to high demands on data collection infrastructure and enormous installation-maintenance costs.  

To mitigate the limitations of the above-mentioned TSE approaches, hybrid TSE methods are introduced, which integrate the traffic knowledge in the form of traffic flow models to ML models for TSE. The hybrid methods based on the learning paradigm of physics-informed deep learning (PIDL) are gaining increasing attentions in recently years,
and is the focus of this paper. 
PIDL contains both a model-driven component (a physics-informed neural network for regularization) and a data-driven component (a physics-uninformed neural network for estimation), making possible the integration of the strengths of both model-driven and data-driven approaches while overcoming the weaknesses of either. 


Despite that the addition of physics could guide the training of PIDL efficiently,  
complicated mathematical formulas could instead make the PIDL difficult to train. There are many theoretical attempts made to add sophistication (usually in the form of complicated terms) to the FD relation for an improved description of the dynamics. 
To balance the sophistication and trainability of encoding physics, this paper explores a promising direction by approximating the FD relation with an ML surrogate, 
instead of hard-encoding an FD equation.
Following this direction, we introduce an improved PIDL paradigm, called physics-informed deep learning with a fundamental diagram learner (PIDL+FDL), which integrates an ML surrogate (e.g., an NN) into the model-driven component to represent the fundamental diagram (FD) and estimate the FD relation.
We focus on highway TSE with observed data from loop detectors, using traffic density or velocity as traffic variables. 
Our contributions are:

\begin{itemize}

\item We propose the PIDL+FDL-based TSE method that possesses advantages to 

\SubItem{Perform the TSE with improved estimation accuracy: A proper integration of  ML surrogates may avoid directly encoding the complicated terms in PIDL and trade off between the sophistication of the model-driven aspect of PIDL and the training flexibility, making the framework a better fit to the TSE problem;}
\SubItem{Perform the FD estimation: The PIDL+FDL uses an ML surrogate to directly learn the underlying FD relation without any FD output measurements, i.e., the ML surrogate is purely trained under physical regularization from PIDL, making it more likely to learn a suitable relation along with the TSE training. It can also get around the calibration of parameters inside the FD equation;}
\SubItem{Perform the model parameter identification: For a complete traffic model reconstruction, in addition to the FD estimation, there may exist model parameters outside the FD terms to be learned. The proposed PIDL+FDL conducts the model parameter identification jointly.}

\item
We validate the PIDL+FDL performance with both numerical experiments and real-world data: 
To demonstrate the strengths of the PIDL+FDL, we design the PIDL+FDL architectures for the traffic dynamics governed by the Greenshields-based LWR and 
ARZ models, respectively. Additionally, experiments using the real-world data, the Next Generation  SIMulation (NGSIM) dataset, are conducted. The experimental results show the advantages of PIDL in terms of estimation accuracy and data efficiency over baselines and the  capacity to properly estimate the FD relation and model parameters.

\end{itemize}

The rest of this paper is organized as follows.  Section II briefs related work on TSE and PIDL. Section III formalizes the PIDL+FDL framework for TSE. Sections IV and V detail the designs and experiments of PIDL+FDL for Greenshields-based LWR and Greenshields-based ARZ, respectively. Section~VI evaluates the PIDL+FDL on NGSIM data over baselines. Section VII concludes our work.

\section{Related Work of Traffic State Estimation}



Most model-driven estimation approaches are data assimilation (DA) based, which find ``the most likely state," allowing observation to correct models' prediction. Popular examples include the Kalman filter (KF) and its variants~\cite{Yibing-2008, xuan-2010}. 
Other than KF-like methods, particle filter (PF)~\cite{Mihaylova-2004} with improved  nonlinear representation, adaptive smoothing filter (ASF)~\cite{Treiber-2011} for combining multiple sensing data, were proposed to improve and extend different aspects of the TSE process.
In data-driven approaches, to handle complicated traffic data, ML approaches were involved, 
including long short term memory (LSTM) and deep embedded models~\cite{LiWei-2018, ZhengZibin-2019}. 

A paradigm integrating physics to ML has gained increasing interests recently. 
Yuan \textit{et al}~\cite{YuanY-2020} proposed a physics regularized Gaussian process
for macroscopic traffic flow modeling and TSE. The hybrid methods using the PIDL framework~\cite{ Raissi-2018b,mo2020physics} is becoming an active field. Huang \textit{et al}~\cite{Huang-Jiheng-2020} studied the use of PIDL to encode the Greenshields-based LWR and validated it in the loop detector scenarios using SUMO simulated data. Barreau \textit{et al}~\cite{Barreau-2020-a, Barreau-2020-b, Barreau-2020-c} studied the probe vehicle sensors and developed coupled micro-macro models for PIDL to perform TSE. Shi \textit{et al}~\cite{Shi-AAAI-2021} extended the PIDL-based TSE to the second-order ARZ with observed data from both loop detectors and probe vehicles. 

We want to highlight that, as to model reconstruction, which is another feature of PIDL-based TSE, this paper only assumes a traffic flow conservation equation and optionally, a momentum equation for the velocity field, without specifying any mathematical relation between traffic quantities. While \cite{Barreau-2020-a, Barreau-2020-c} directly fit a velocity function using measured density and velocity from probe vehicles before or during the PIDL training, we, in this paper, focus on a more general case, where the output of the FD function is unobserved from sensors, and the end-to-end FD relation is learned directly using ML surrogates under the PIDL framework.
In summary, this paper contributes to the trend of developing hybrid methods for TSE and model reconstruction, especially with both FD estimation and model parameter identification involved. 

\section{Mathematical Setting for PIDL+FDL}

This section introduces the PIDL+FDL framework in the context of TSE at a high level.

\subsection{PIDL for TSE}
\label{3-A-r3}

Consider a traffic flow dynamics of a road segment that is governed by a set of non-linear equations (e.g., partial differential equations, PDEs):

\begin{equation}
\label{equ-3-1-r3}
\mathcal{N}(\pmb{M}(t,x), \pmb{Q};\pmb{\lambda}) = \pmb{0}, x\in [0,L], t \in [0,T],
\end{equation}

\noindent
where $L\in \mathbb{R}^+$, $T\in \mathbb{R}^+$. We use bold symbols to denote vectors by default. The operator $\mathcal{N}$ contains the governing non-linear equations of the traffic flow dynamics, while $\pmb{M}(t,x)$ contains the traffic state variables, such as the traffic density $\rho(t,x)$ and velocity $u(t,x)$. $\pmb{\lambda}$ contains the model parameters. The model includes intermediate unobserved traffic variables $\pmb{Q}$ that have some hidden relationship with $\pmb{M}(t,x)$. Thus, the dynamics can be represented by

\begin{equation}
\label{equ-3-2-r3}
\mathcal{N}(\pmb{M}, \pmb{Q}(\pmb{M});\pmb{\lambda}) = \pmb{0},
\end{equation}

\noindent
and $\pmb{M}$ stands for $\pmb{M}(t,x)$. For general discussion, the values of $\pmb{Q}$ are not assumed to be directly observable, and the relation is either unknown or deduced based on assumptions which may be deficient. The TSE problem is to reconstruct the traffic states $\pmb{M}$ at each point $(t,x)$ in a continuous domain from partial observation of $\pmb{M}$. Accordingly, the  continuous spatio-temporal domain $D$ is a set of points: $ D=\{(t,x)| \forall t\in [0,T], x\in [0,L] \}$. We represent this continuous domain in a discrete manner using grid points $G \in D$ that are evenly deployed throughout the domain. We define the set of grid points as $G=\{(t^{(r)},x^{(r)})|r=1,..,N_g  \}$. The total number  of grid points, $N_g$, controls the fine-grained level of $G$ as a representation of the continuous domain.

PIDL approximates $\pmb{M}(t,x)$ using a neural network with time~$t $ and location $x$ as its inputs. This neural network is called \textit{physics-uninformed neural network} (PUNN) (or \textit{estimation network} in our TSE study), which is parameterized by~$\theta$. We denote the approximation of  $\pmb{M}(t,x)$ from PUNN as $\hat{\pmb{M}}(t,x;\theta)$. When $\mathcal{N}$, $\pmb{Q}$ and $\pmb{\lambda}$ are known, during the learning phase of PUNN (i.e., to find the optimal $\theta$ for PUNN), the following equiation defines the residual values of the approximation $\hat{\pmb{M}}(t,x;\theta)$:

\begin{equation}
\label{equ-3-3-r3}
\pmb{\hat{f}}(t,x;\theta):=\mathcal{N}(\hat{\pmb{M}}(t,x;\theta), \pmb{Q}(\hat{\pmb{M}}(t,x;\theta));\pmb{\lambda}),
\end{equation}

\noindent which is designed according to the traffic flow model in Eq.~(\ref{equ-3-2-r3}). The calculation of residual $\pmb{\hat{f}}(t,x;\theta)$ is done by a \textit{physics-informed neural network} (PINN). This network can compute $\pmb{\hat{f}}(t,x;\theta)$ directly using  $\hat{\pmb{M}}(t,x;\theta)$, the output of PUNN, as its input. When  $\hat{\pmb{M}}(t,x;\theta)$ is closer to the true value $\pmb{M}(t,x)$, the residual $\pmb{\hat{f}}$ will be closer to $\pmb{0}$. PINN introduces no new parameters, and thus, shares the same $\theta$ of PUNN.

In most cases, even the model is given, the model parameters $\pmb{\lambda}$  are unknown and can be made as learning variables in PINN for model parameter identification. The residual $\hat{\pmb{f}}$ is then redefined as 

\begin{equation}
\label{equ-3-4-r3}
\pmb{\hat{f}}(t,x;\theta,\pmb{\lambda}):=\mathcal{N}(\hat{\pmb{M}}(t,x;\theta), \pmb{Q}(\hat{\pmb{M}}(t,x;\theta));\pmb{\lambda}).
\end{equation}

\noindent
This paper assumes unknown model parameters by default.

In PINN, $\pmb{\hat{f}}(t,x;\theta,\pmb{\lambda})$ is calculated by automatic differentiation technique, which can be done by the function {\ttfamily \footnotesize tf.gradient} in Tensorflow. The activation functions and the connecting structure of neurons in PINN are designed to conduct the differential operation in Eq.~(\ref{equ-3-4-r3}). We would like to emphasize that, the connecting weights in PINN have fixed values which are determined by the traffic flow model and $\pmb{\lambda}$ are encoded as learning variables. Thus, the residual $\pmb{\hat{f}}$ is parameterized by both $\theta$ and $\pmb{\lambda}$.

The training data for PIDL consist of (1) \textit{observation points} $O=\{(t^{(i)}_o, x^{(i)}_o) | i=1,...,N_o\}$, (2) \textit{target values} $P=\{\pmb{M}^{(i)}| i=1,...,N_o\}$ (i.e., the true traffic states at the observation points), and (3) \textit{auxiliary points} $A=\{(t^{(j)}_a, x^{(j)}_a)| j=1,...,N_a\}$. $i$ and $j$ are the indexes of observation points and auxiliary points, respectively. One target value is associated with one observation point, and thus, $O$ and $P$ have the same indexing system (indexed by $i$). This paper uses the term, observed data, to denote $\{O,P\}$. Both $O$ and $A$ are subsets of the grid points $G$ (i.e., $O \in G$ and $A \in G$). 

Observation points $O$ are limited to the time and locations that traffic sensors can visit and record. In contrast, auxiliary points $A$ have neither measurement requirements nor location limitations, and the number of $A$ is controllable. $A$ are used for regularization purposes, and this is why they are called ``auxiliary''. To train a PUNN for TSE, the loss is used:

\begin{equation}
\label{equ-3-5-r3}
\begin{split}
\begin{gathered}
Loss_{\theta,\pmb{\lambda}}=\alpha \cdot MSE_o + \beta \cdot MSE_a \\
= \alpha \cdot \underbrace{\frac{1}{N_o} \sum\limits_{i=1}^{N_o} |\hat{\pmb{M}}(t^{(i)}_o, x^{(i)}_o;\theta)-\pmb{M}^{(i)}|^2}_{data\  discrepancy} \\ +   \beta \cdot \underbrace{\frac{1}{N_a}\sum\limits_{j=1}^{N_a} |{\pmb{\hat{f}}}(t^{(j)}_a, x^{(j)}_a;\theta, \pmb{\lambda})|^2}_{physical\  discrepancy}.
\end{gathered}
\end{split}
\end{equation}

\noindent where $\alpha$ and $\beta$ are  hyperparameters for balancing the contribution  to the loss made by data discrepancy and physical discrepancy, respectively. The data discrepancy is defined as the mean square error (MSE) between approximation $\hat{\pmb{M}}$ on $O$ and target values $P$. The physical discrepancy is the MSE between residual values on $A$ and $\pmb{0}$, quantifying the extent to which the approximation deviates from the traffic model.

Given the training data, we apply neural network training algorithms to solve  $(\theta^*, \pmb{\lambda}^*) = \mathrm{argmin}_{\theta,\pmb{\lambda}}\  Loss_{\theta, \pmb{\lambda}}$. Then, the $\pmb{\lambda}^*$-parameterized traffic flow model of Eq.~(\ref{equ-3-4-r3}) is the most likely physics that generates the observed data, and the $\theta^*$-parameterized PUNN can then be used to approximate the traffic states on $G$, which are consistent with the reconstructed traffic flow model in Eq.~(\ref{equ-3-2-r3}).

\subsection{PIDL + FDL for TSE}
\label{3-B-r3}

As has been discussed previously, the PIDL-based TSE methods may perform poorly when informed by a highly sophisticated traffic flow model. This is because the models may contain complicated terms that are unfriendly to differentiation-based learning (e.g., square root operators of learning variables in the denominator, etc.), making the training and performance very sensitive to the structural design of PINN. Many efforts such as variable conversion, decomposition and factorization need to be made to have the PINN trainable and the loss to converge. In our framework of Eq.(\ref{equ-3-2-r3}), these ``unfriendly'' terms can be contained as part of the hidden relation $\pmb{Q}$. To address the issues of PIDL-based TSE, we propose to use an ML surrogate $\pmb{\hat{Q}}$ to directly represent the  $\pmb{Q}$ and learn the relation under the PIDL framework, instead of hard-encoding a complicated term in PINN.

The advantages of properly introducing an ML surrogate of $\pmb{Q}$ are two-fold: (1) An ML term is usually differentiation-friendly, giving the PIDL more flexibility to achieve an improved TSE accuracy. (2) No assumptions are made to the hidden relation, and it is possible to learn a more suitable $\pmb{\hat{Q}}$ when trained under the physical regularization from PINN.

This paper focuses on one kind of hidden relationships, the fundamental diagram (FD), and the corresponding learning paradigm is called \textit{PIDL with an FD Learner} (PIDL+FDL). Specifically, the FD Learner, formalized as $\pmb{\hat{Q}}(\pmb{\hat{M}};\omega)$, can be designed as  a neural network parameterized by $\omega$ to represent the unknown FD relation, which takes the estimated traffic variables $\pmb{\hat{M}}(t,x;\theta)$ as its input. The residual of Eq.(\ref{equ-3-4-r3}) is redefined as the following

\begin{equation}
\label{equ-3-6-r3}
\pmb{\hat{f}}(t,x;\theta,\omega,\pmb{\lambda}):=\mathcal{N}(\hat{\pmb{M}}(t,x;\theta), \hat{\pmb{Q}}(\hat{\pmb{M}}(t,x;\theta);\omega);\pmb{\lambda}).
\end{equation}

\noindent
The loss function becomes

\begin{equation}
\label{equ-3-7-r3}
\begin{split}
\begin{gathered}
Loss_{\theta,\omega,\pmb{\lambda}}=\alpha \cdot MSE_o + \beta \cdot MSE_a \\
= \alpha \cdot \frac{1}{N_o} \sum\limits_{i=1}^{N_o} |\hat{\pmb{M}}(t^{(i)}_o, x^{(i)}_o;\theta)-\pmb{M}^{(i)}|^2 \\ +   \beta \cdot \frac{1}{N_a}\sum\limits_{j=1}^{N_a} |{\pmb{\hat{f}}}(t^{(j)}_a, x^{(j)}_a;\theta,\omega, \pmb{\lambda})|^2.
\end{gathered}
\end{split}
\end{equation}

\noindent
Using the training data, we apply neural network training algorithms to solve   $(\theta^*,\omega^*, \pmb{\lambda}^*) = \mathrm{argmin}_{\theta,\omega,\pmb{\lambda}}\  Loss_{\theta,\omega, \pmb{\lambda}}$.  Then, in addition to TSE learning and model parameter identification, the FD estimation is conducted automatically. The $\omega^*$-parameterized $\pmb{\hat{Q}}$ can be used to represent the unknown hidden fundamental diagram relation. Note that the values of $\pmb{Q}$ are not assumed to be observable, and thus, are not part of the data (i.e., to directly learn the $\pmb{\hat{Q}}$ from data cannot apply).

In some cases, the curve of the learned $\pmb{\hat{Q}}$ may present abnormal shapes on edge conditions. To mitigate this, one can encode prior knowledge into the loss as an additional regularization term $Reg(\pmb{\hat{Q}})$ to reshape the FD. As an example, we can use $\pmb{Q}$ to represent the density-flow relation, i.e., the flux function (one typical kind of FD), mapping the density $\rho$ to the flow value, which is denoted as a \textit{scalar} $Q(\rho)$. Existing theoretical works usually assume $Q$ to be concave with respect to the traffic density $\rho$. To impose the concavity property, we design the following regularization term:

\begin{equation}
\label{equ-3-8-r3}
Reg(Q)=\int_a^b {\max (0,\frac{{\partial ^2 Q(\rho )}}{{\partial \rho ^2 }})d\rho } ,
\end{equation}

\noindent
where the hyperparameters $a$ and $b$ determine the interval of $\rho$ on which the reshaping takes effects without interfering the learning on other regions. We propose this design because most abnormal shapes only occur on edge region and it is not necessary to regularize over the whole traffic density domain. We apply $Loss_{\theta,\omega,\pmb{\lambda}}=\alpha \cdot MSE_o + \beta \cdot MSE_a+\xi \cdot Reg(\hat{Q})$ in the learning phase and properly reshape the learned FD curves.


\section{PIDL+FDL for Greenshields-Based LWR}
\label{sec-IV}

The first numerical example aims to show the capability of our method to estimate the traffic dynamics governed by the LWR model with a Greenshields flux function.

Define flow rate $Q$ (a scalar) to be the number of vehicles passing a specific position on the road per unit time, and traffic density $\rho$ to be the average number of vehicles per unit length of the road.  The traffic flux $Q(\rho)$ describes  $Q$ as a function of  $\rho$, which is the FD relation of interest in this numerical example. We treat $\rho$ as the basic traffic state variable to estimate. Greenshields flux is a basic and popular choice of $Q(\rho)$, which is defined as $Q(\rho)=\rho u_{max}(1-\rho/\rho_{max})$, where $u_{max}$ and $\rho_{max}$ are maximum velocity and maximum (jam) density, respectively. This flux function has a quadratic form with two coefficients $u_{max}$ and $\rho_{max}$.

The LWR model~\cite{Lighthill-1955, Richards-1956} describes the macroscopic traffic flow dynamics as $\rho_t+(Q(\rho))_x=0$, which is derived from a conservation law of vehicles. In order to reproduce more phenomena in observed traffic data, such as delayed driving behaviors due to drivers' reaction time, diffusively corrected LWRs were introduced, by adding a diffusion term, containing a second-order derivative $\rho_{xx}$. 
We focus on one version of the diffusively corrected LWRs:  $\rho_t+(Q(\rho))_x=\epsilon \rho_{xx}$, where $\epsilon$ is the diffusion coefficient.

In this section, we study the Greenshields-based LWR traffic flow model of a ``ring road":

\begin{equation}
\label{equ-3-9-r3}
\begin{split}
\begin{gathered}
\rho_t + (Q(\rho))_x=\epsilon \rho_{xx}, \  t\in [0,3], \  x\in [0,1],\\
Q(\rho)=\rho \cdot u_{max}\Bigl( 1- \frac{\rho}{\rho_{max}} \Bigl) \ \ \ (FD\ relation), \\
\rho(t,0)=\rho(t,1) \ \ \ (boundary\ condition\ 1), \\
\rho_x(t,0)=\rho_x(t,1) \ \ \ (boundary\ condition\ 2),
\end{gathered}
\end{split}
\end{equation}

\noindent
where $\rho_{max}=1.0$, $u_{max}=1.0$, and $\epsilon = 0.005 $. $\rho_{max}$ and $u_{max}$ are usually determined by physical restrictions of the road and vehicles.

Given the bell-shaped initial $0.1+0.8e^{-(\frac{x-0.5}{0.2})^2}$, $x \in [0,1]$, we apply the Godunov scheme to solve Eqs.~(\ref{equ-3-9-r3}) on 960 (time) $\times$ 240 (space) grid points $G$ evenly deployed throughout the $[0,3]\times [0,1]$  domain. In this case, the total number of grid points $G$ is $N_g=$960$\times$240. The numerical solution is shown in Fig.~\ref{fig:deployment-r3} (see the heat map background). From the figure, we can visualize the dynamics as follows: At $t=0$, there is a peak density at the center of the road segment, and this peak evolves to propagate along the direction of $x$, which is known as the phenomenon of  traffic shockwave. Since this is a ring road, the shockwave reaching $x=1$ continues at $x=0$. This numerical solution of the Greenshields-based LWR model is treated as the ground-truth traffic density. We will apply a PIDL+FDL-based approach method to estimate the entire traffic density field using observed data as well as to estimate the FD relation and model parameters.


%

\subsection{PIDL+FDL Architecture Design}

The authors' previous work~\cite{Shi-AAAI-2021} has shown the capacity of PIDL to perform both TSE and model parameter identification when the closed traffic flow model is given. Here we are only given the knowledge of conservation law and boundary conditions, i.e., the FD relation is unknown and no direct observation of $Q$ is available. 

We employ a neural network $\hat{Q}(\cdot;\omega)$ to estimate the traffic flow from the traffic density and to represent the FD relation of interest. Based on Eqs.~(\ref{equ-3-9-r3}), we define the residual value of PUNN's traffic density estimation $\hat{\rho}(t,x;\theta)$ as

\begin{equation}
\label{equ-10-r3}
\hat{f}(t,x;\theta,\omega,\epsilon) := \hat{\rho}_t(t,x;\theta) + (\hat{Q}(\hat{\rho}(t,x;\theta);\omega))_x-\epsilon \hat{\rho}_{xx}(t,x;\theta).
\end{equation}

\noindent
Note that the parameter $\pmb{\lambda}$ contains the coefficient $\epsilon$ only.

\begin{figure}[t!]
\centering
  \includegraphics[scale=0.62]{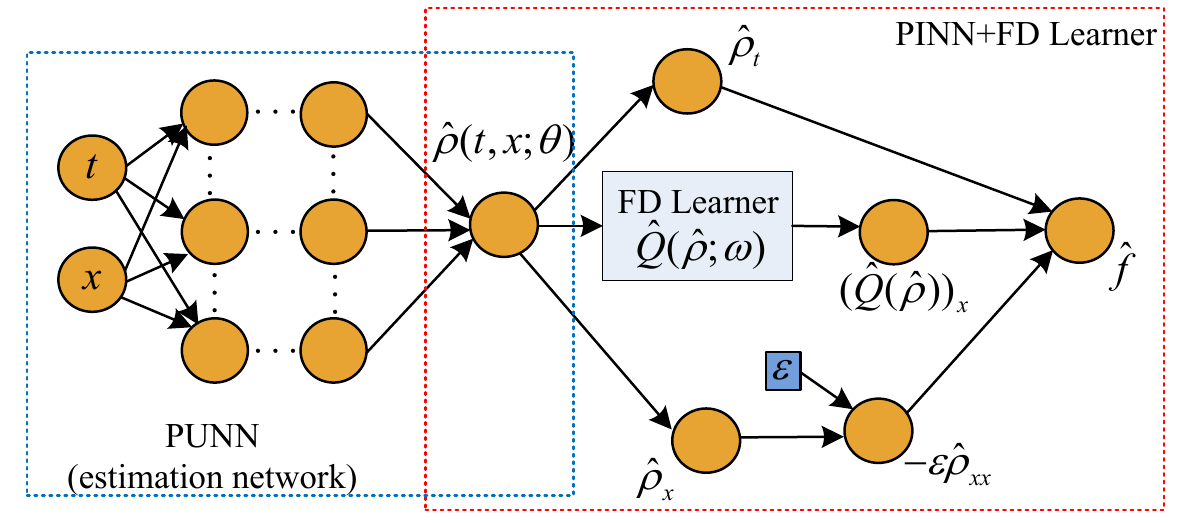}
  \caption{PIDL+FDL architecture for first-order traffic dynamics, consisting of a PUNN for traffic density estimation and a PINN+FD Learner for calculating the residual Eq.~(\ref{equ-10-r3}). The model parameter $\epsilon$ is held by a variable node (blue rectangular nodes). All connecting weights are predefined and fixed in PINN.}
  \label{fig:ch4-PINN_structure-r3}
\end{figure}

Given the definition of $\hat{f}$, the corresponding PIDL+FDL architecture is shown in Fig.~\ref{fig:ch4-PINN_structure-r3}. This architecture consists of a PUNN for traffic density estimation, followed by a PINN+FD Learner for calculating the residual Eq.~(\ref{equ-10-r3}). The PUNN parameterized by $\theta$ is designed as a fully-connected feedforward neural network with 8 hidden layers and 20 hidden nodes in each hidden layer. Hyperbolic tangent function (tanh)  is used as the activation function for each hidden neuron in PUNN. In contrast, in PINN, connecting weights are fixed and the activation function of each node is designed to conduct specific nonlinear operation for calculating an intermediate (hidden) value of~$\hat{f}$. The flow value is calculated by a separate neural network $\hat{Q}(\hat{\rho};\omega)$ with two hidden layers and 20 hidden nodes for each. The model parameter $\epsilon$ is held by a variable node (blue rectangular nodes).

To customize the training of PIDL+FDL to  Eqs.~(\ref{equ-3-9-r3}), in addition to the training data $O$, $P$ and $A$ defined in Section~\ref{3-A-r3}, we  need to introduce \textit{ boundary auxiliary points} $B=\{(t^{(k)}_b,0)| k = 1,...,N_b\} \cup \{(t^{(k)}_b,1)| k = 1,...,N_b\}$, for learning the two boundary conditions in Eqs.~(\ref{equ-3-9-r3}).

For experiments of state estimation with both parameter identification and FD estimation, we design the following loss

\begin{equation}
\label{equ-11-r3}
\begin{split}
Loss_{\theta,\omega,\epsilon} &= \alpha \cdot MSE_o +  \beta \cdot MSE_a + \gamma \cdot B1 + \eta \cdot B2 \\
&=  \frac{\alpha}{N_o} \sum\limits_{i=1}^{N_o} |\hat{\rho}(t^{(i)}_o, x^{(i)}_o;\theta)-\rho^{(i)}|^2 \\
&  +    \frac{\beta}{N_a}\sum\limits_{j=1}^{N_a} |\hat{f}(t^{(j)}_a, x^{(j)}_a;\theta,\omega,\epsilon)|^2 \\
&+ \frac{\gamma}{N_b} \sum\limits_{k=1}^{N_b} |\hat{\rho}(t^{(k)}_b, 0;\theta)-\hat{\rho}(t^{(k)}_b, 1;\theta)|^2 \\
&+ \frac{\eta}{N_b} \sum\limits_{k=1}^{N_b} |\hat{\rho}_x(t^{(k)}_b, 0;\theta)-\hat{\rho}_x(t^{(k)}_b, 1;\theta)|^2 .
\end{split}
\end{equation}

\noindent  $B1$, scaled by $\gamma$, is the MSE between estimations at the two boundaries $x=0$ and $x=1$. $B2$, scaled by $\eta$, quantifies the difference of first order derivatives at the two boundaries.

\subsection{TSE+FDL using Observation from Loop Detectors}
\label{IV-B}

\begin{figure}[h!]
\centering
  \includegraphics[scale=0.55]{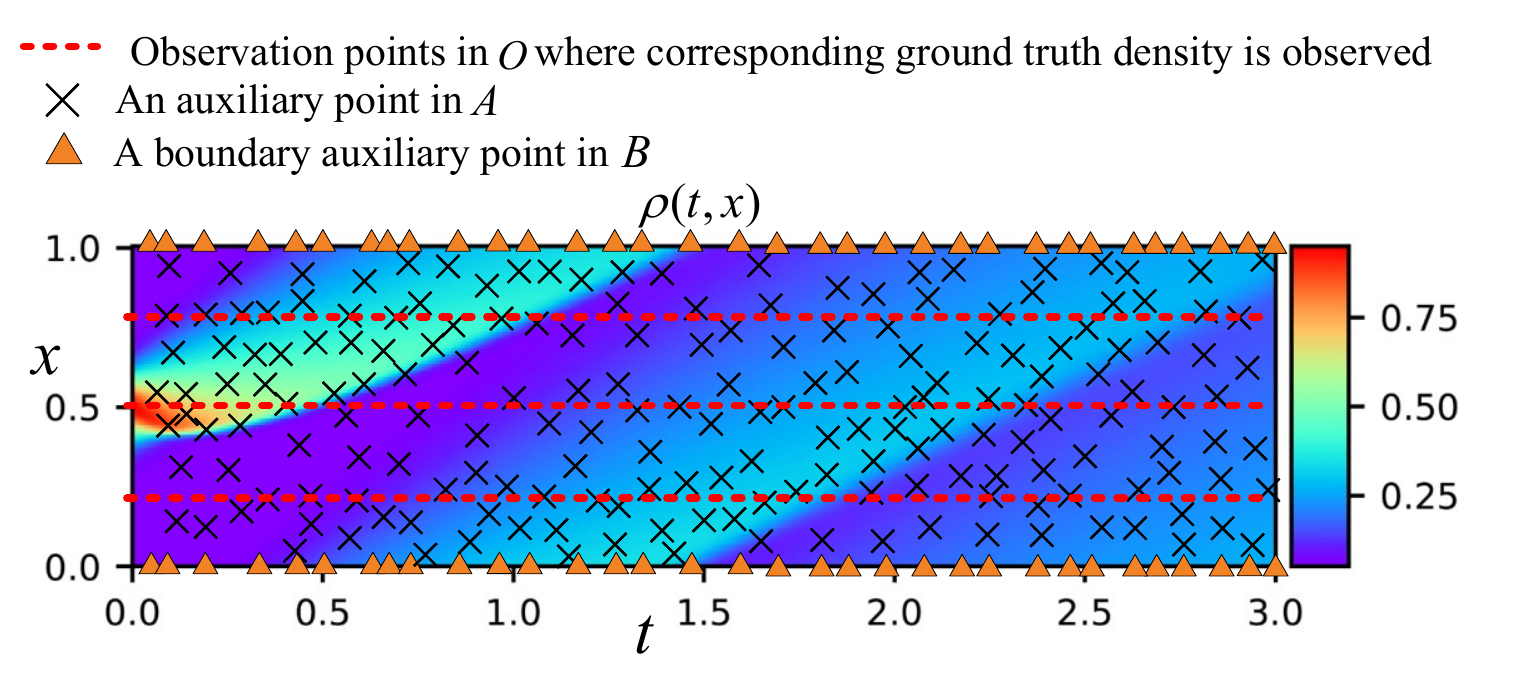}
  \caption{A sparse presentation of the deployment of observation points $O$ at loop detectors, auxiliary points $A$ randomly selected from $G$, and boundary auxiliary points $B$ deployed  at the boundaries $x=0$ and $x=1$ for certain time points. The heatmap is the numerical solution of Eqs.~(\ref{equ-3-9-r3}) using the Godunov scheme. We treat the numerical solution as the ground truth.}
  \label{fig:deployment-r3}
\end{figure}

We justify the capacity of PIDL+FDL in Fig.~\ref{fig:ch4-PINN_structure-r3} for estimating the traffic density field using observation from loop detectors, i.e., only the traffic density at certain locations where loop detectors are installed can be observed. By default, loop detectors are evenly located along the road. To be specific, the grid points at certain locations are used as the observation points $O$, and their corresponding densities constitute the target values $P$ for training. There are $N_a= 100,000$ auxiliary points in $A$ randomly selected from grid points $G$. $N_b=650$ out of 960 \textit{grid time points} (i.e., the time points on the temporal dimension of $G$) are randomly selected to create boundary  auxiliary  points $B$. A sparse version of the deployment of $O$, $A$ and $B$ in the spatio-temporal domain is shown in Fig.~\ref{fig:deployment-r3}. Each observation point is associated with a target value in $P$. Note $O$, $A$ and $B$ are all subsets of $G$.

We train the proposed PIDL+FDL on an NVIDIA Titan RTX GPU with 24 GB memory. By default, we use the $\mathbbm{L}^2$ relative error on $G$ to quantify the estimation error of the entire domain:

\begin{equation}
\label{equ-12-r3}
Err(\hat{\rho},\rho)= \frac{\sqrt{\sum_{r=1}^{N_g} \bigl|\hat{\rho}(t^{(r)}, x^{(r)};\theta)-\rho(t^{(r)}, x^{(r)}) \bigl|^2}}{\sqrt{\sum_{r=1}^{N_g} \bigl|\rho(t^{(r)}, x^{(r)}) \bigl|^2}}.
\end{equation}

\noindent
The reason for choosing the $\mathbbm{L}^2$ relative error is to normalize the estimation inaccuracy, mitigating the influence from the scale of true density values. One remark is that there are some TSE methods (e.g., non parametric ones) that do not perform any estimation on the observation points and directly use the target values there. For these cases, the observation points will be removed from $G$ before calculating Eq.(\ref{equ-12-r3}).

We use the Xavier uniform initializer to initialize $\theta$ of PUNN and $\omega$ of FD Learner (FDL). This neural network initialization method takes the number of a layer's incoming and outgoing network connections into account when initializing the weights of that layer, which may lead to a good convergence. The $\epsilon$ is initialized at 0. Then, we train the PUNN, FDL and $\epsilon$ through the PIDL+FDL architecture using a popular stochastic gradient descent algorithm, the Adam optimizer, for a rough training. A follow-up fine-grained training is done by the L-BFGS optimizer~\cite{Byrd-1995} for stabilizing the convergence, and the process terminates until the loss change of two consecutive steps is no larger than $10^{-16}$. This training process converges to a local optimum $\theta^*$, $\omega^*$ and $\epsilon^*$ that minimize the loss in Eq.~(\ref{equ-11-r3}).

We would like to clarify that in this paper, the training data are the observed data from detectors, i.e., the traffic states on the points at certain locations where loops are equipped. 

\begin{figure}[h!]
\centering
  \includegraphics[scale=0.5]{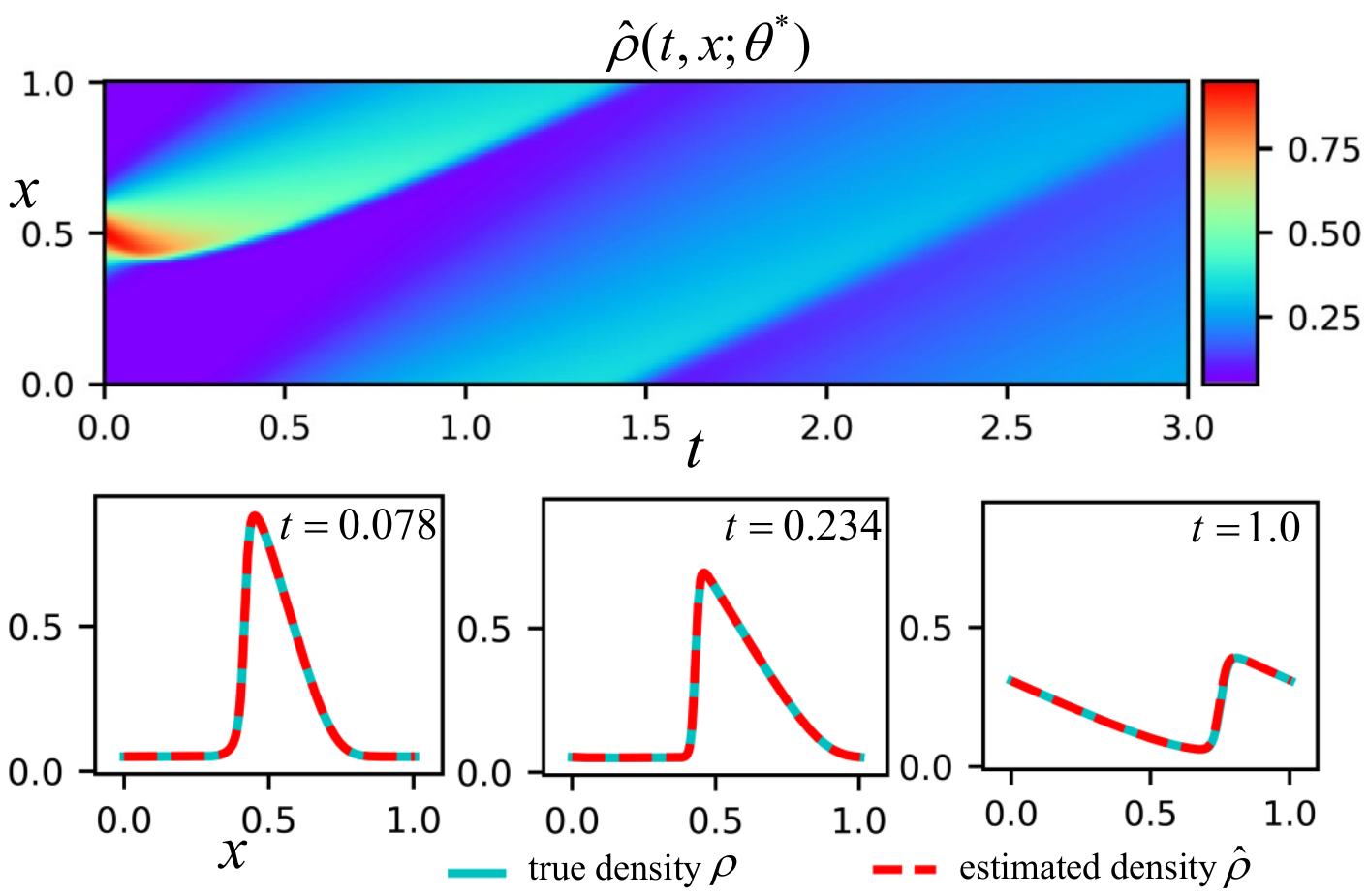}
  \caption{Top: Estimation of the traffic density dynamics $\hat{\rho}(t,x;\theta^*)$ on  grid points $G$ in the domain using the trained PUNN. Bottom: Snapshots  of  estimated and true traffic density at certain time points.}
  \label{fig:ch4-results-r3}
\end{figure}

The results of applying the PIDL+FDL with 4 loops to the Greenshields-based LWR dynamics is presented in Fig.~\ref{fig:ch4-results-r3}, where PUNN is parameterized by the optimal $\theta^*$. As shown in Fig.~\ref{fig:ch4-results-r3}, the estimation $\hat{\rho}(t,x;\theta^*)$ is visually the same as the true dynamics $\rho(t,x)$ in Fig.~\ref{fig:deployment-r3}. By looking into the estimated and true traffic density over $x$ at certain time points, there is a good agreement between two traffic density curves. The $\mathbbm{L}^2$ estimation error  $Err(\hat{\rho},\rho)$ is $1.287\times 10^{-2}$. Empirically, the difference cannot be visually distinguished when the estimation error is smaller than $6\times 10^{-2}$.

\begin{table}[h!]
    \caption{Performance of PIDL+FDL on the LWR Dynamics}
    \label{tab:ch4-r3}
    \centering
    \begin{threeparttable}
    \begin{tabular}{ccccc}
    \toprule
    \toprule
    \multicolumn{1}{c}{$loops$} & \multicolumn{1}{c}{2} & \multicolumn{1}{c}{3} & \multicolumn{1}{c}{4} &
    \multicolumn{1}{c}{5}\\
    \midrule
    $Err(\hat{\rho},\rho)$    & 0.6021 & 0.03327 & 0.01287 & 0.004646\\ 
    $\epsilon^*$    & 3.12573 & 0.00495 & 0.00506 & 0.00509\\
    \bottomrule
    \bottomrule
    \end{tabular}
    \begin{tablenotes}
      \footnotesize
      \item $loops$ stands for the number of loop detectors.  $\epsilon^*$ is the estimated diffusion coefficient. Note the true $\epsilon=0.005$.
    \end{tablenotes}
  \end{threeparttable}
\end{table}

We change the number of loop detectors. For a fixed number of loop detectors, we use grid search for hyperparameter tuning by default. Specifically, since Adam optimizer is scale invariant, we fix the hyperparameter $\alpha$ to 100 and tune the other hyperparameters from [1, 10, 50, 100, 150, 200] with some follow-up fine tuning. The minimal-achievable estimation errors of PIDL+FDL over the numbers of loop detectors are presented in Table~\ref{tab:ch4-r3}. From the table, we can see that the traffic density estimation errors improve as the number of loop detectors increases. When more than two loop detectors are used, the model parameters to be learned are able to converge to the true parameters $\epsilon$. Specifically, with three loop detectors, in addition to a good traffic density estimation error of 3.327$\times 10^{-2}$, the model parameter converges to $\epsilon^*=0.00495$, which is very close to the true value 0.005. The results demonstrate that PIDL+FDL method can handle both TSE and model parameter identification with three loop detectors for the traffic dynamics of the Greenshields-based LWR.


In this experiment, the proposed method can reconstruct the exact Greenshields FD when 5 loop detectors are used. The results are meaningful because neither any assumptions on the FD relation are made nor the flow values are observed directly. In addition to traffic density estimation and model parameter identification, the PIDL+FDL is able to make full use of the conservation law and boundary conditions to retrieve the density-flow relation automatically. For the visualization of the results, please refer to Figure 4 in the supplementary material in~\cite{Shi-arXiv-2021}.

\section{PIDL+FDL for Greenshields-Based ARZ}
\label{sec-V-r3}

The second numerical example aims to show the capacity of the proposed method to handle the traffic dynamics governed by the Greenshields-based ARZ, a second-order traffic flow model with both traffic density $\rho$ and velocity $u$  as the traffic state variables.

An ARZ model involves both a conservation law of vehicles and a momentum equation on velocity. Specifically, we study the following traffic flow dynamics of a ``ring road” in  $t\in [0,3]$, $x\in [0,1]$:

\begin{equation}
\label{equ-4-13-r3}
\begin{split}
\begin{gathered}
\rho_t + (\rho u)_x= 0,\\
(u+h(\rho))_t+u(u+h(\rho))_x= (U_{eq}(\rho) - u)/\tau , \\
h(\rho)=U_{eq}(0)- U_{eq}(\rho) \ \ \  (traffic\ \ pressure), \\
U_{eq}(\rho)=  u_{max}(1-\rho/\rho_{max})  \ \ \  (equilibrium\ \ speed),\\
\rho(t,0)=\rho(t,1), u(t,0)=u(t,1) \  (boundary\ cond.),
\end{gathered}
\end{split}
\end{equation}

\noindent where we set the parameters irregularly as $\rho_{max}=1.13$, $u_{max}=1.02$, and $\tau = 0.02 $. $U_{eq}$ is the equilibrium velocity, $h(\rho)$ defines the traffic pressure and $\tau$ denotes the relaxation time scale. For more explanations of this ARZ setting, we refer readers to our previous work in \cite{Shi-AAAI-2021}.

\begin{figure}[h!]
\centering
  \includegraphics[scale=0.82]{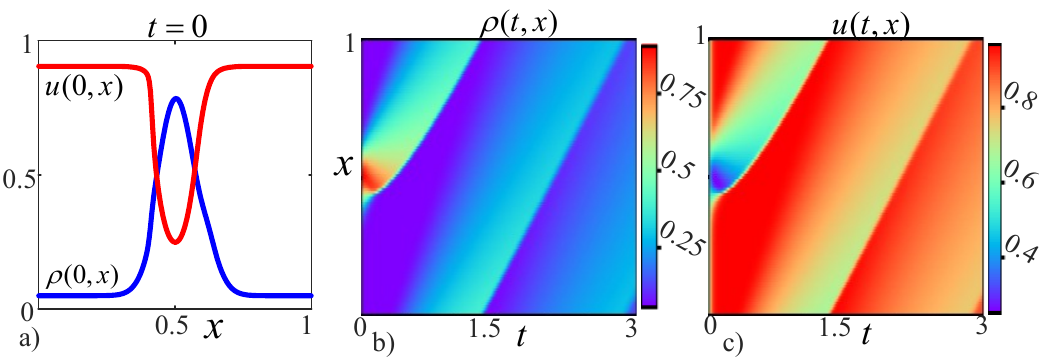}
  \caption{a) is the bell-shaped initial $\rho$ and $u$ over $x\in [0,1]$; b) and c) are numerical solutions for $\rho$ and $u$, respectively.}
  \label{fig:ch5-initial-r3}
\end{figure}

Given the bell-shaped initial of $\rho$ and $u$ as shown in Fig.\ref{fig:ch5-initial-r3}.a, we apply the Lax–Friedrichs (LF) scheme to solve Eqs.(\ref{equ-4-13-r3}) on grid $G$ with 960 (time)$\times$240 (space) points evenly deployed over the $[0,3]\times [0,1]$ domain. The LF numerical solutions of both $\rho$ and $u$ over the domain are shown in Fig.\ref{fig:ch5-initial-r3} as well. We treat this numerical solution as the ground-truth to test our PIDL+FDL-based approach for the ARZ dynamics.

\subsection{PIDL+FDL Architecture Design for ARZ}

We employ a neural network $\hat{U}_{eq}(\cdot;\omega)$ to estimate the equilibrium velocity and $U_{eq}(\rho)$ is the target FD relation in this numerical example. Based on Eqs.~(\ref{equ-4-13-r3}), we define the following residuals

\begin{equation}
\label{equ-8}
\begin{split}
  {\begin{array}{l}
      \hat{f}_1(t,x;\theta):=\hat{\rho}_t + (\hat{\rho} \hat{u})_x ,  \\
    \hat{f}_2(t,x;\theta,\omega,\tau):=(\hat{u}+h(\hat{\rho}))_t+\hat{u}(\hat{u}+h(\hat{\rho}))_x \\  \ \ \ \ \ \ \ \ \ \ \ \ \ \ \ \ \ \ \ \ - (\hat{U}_{eq}(\hat{\rho};\omega) - \hat{u})/\tau ,
\end{array}}
\end{split}
\end{equation}

\noindent where $\hat{\rho}$ and $\hat{u}$ are shorthands for $\hat{\rho}(t,x;\theta)$ and $\hat{u}(t,x;\theta)$, respectively outputted from a PUNN. By using the FD surrogate $\hat{U}_{eq}$, the relaxation $\tau$ is the only model parameter to be learned.

\begin{figure}[h!]
\centering
  \includegraphics[scale=0.7]{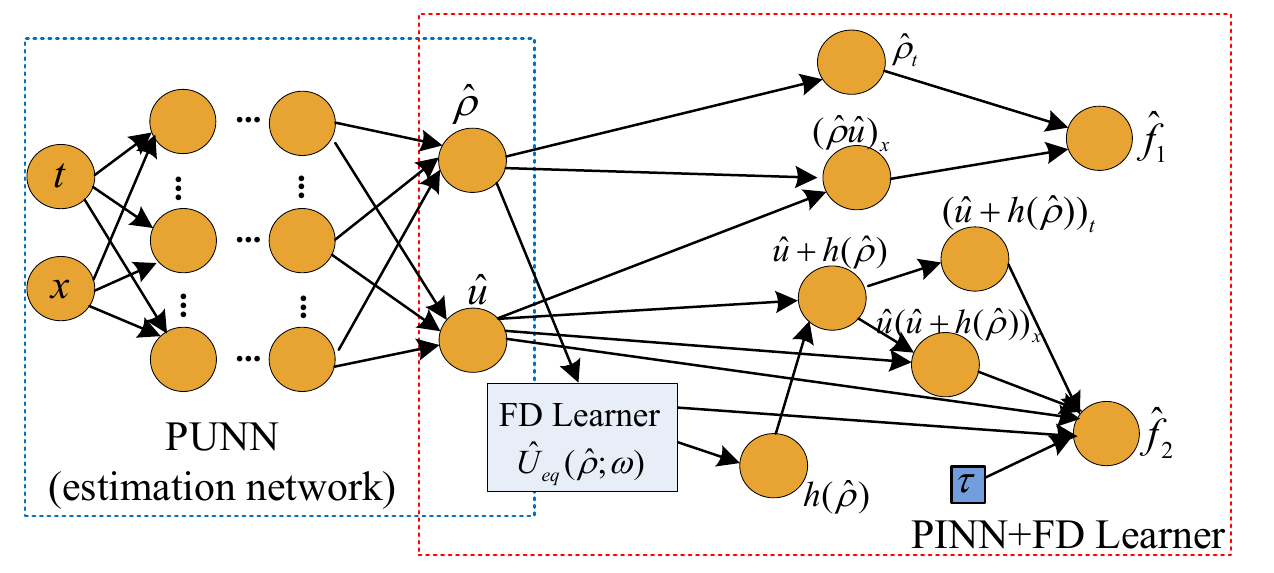}
  \caption{PIDL+FDL architecture for second-order traffic dynamics. The model parameter $\tau$ is held by a variable node (blue rectangular node).}
  \label{fig:ch4-ARZ-PIDL-r3}
\end{figure}

Given the definition of $(\hat{f}_1$, $\hat{f}_2)$, the design of PINN+FDL architecture is shown in Fig.~\ref{fig:ch4-ARZ-PIDL-r3}. The structures of hidden layers of the PUNN and FDL are the same with those of Section~\ref{sec-IV}. For this experiment, we adjust the learning loss as the following:

\begin{equation}
\label{equ-15-r3}
\begin{split}
\begin{gathered}
Loss_{\theta,\omega,\tau}=  MSE_o + MSE_a +B1 \\  =  \frac{1}{N_o} \sum\limits_{i=1}^{N_o}  \alpha_1|\hat{\rho}(t^{(i)}_o, x^{(i)}_o;\theta)-\rho^{(i)}|^2 + \alpha_2|\hat{u}(t^{(i)}_o, x^{(i)}_o;\theta)-u^{(i)}|^2 \\       +\frac{1}{N_a}\sum\limits_{j=1}^{N_a} \beta_1|\hat{f_1}(t^{(j)}_a, x^{(j)}_a;\theta)|^2+\beta_2|\hat{f}_2(t^{(j)}_a, x^{(j)}_a;\theta,\omega,\tau)|^2 \\
  + \frac{1}{N_b} \sum\limits_{k=1}^{N_b} (\gamma_1 |\hat{\rho}(t^{(k)}_b, 0;\theta)-\hat{\rho}(t^{(k)}_b, 1;\theta)|^2 \\   + \gamma_2 |\hat{u}(t^{(k)}_b, 0;\theta)-\hat{u}(t^{(k)}_b, 1;\theta)|^2 ).
\end{gathered}
\end{split}
\end{equation}

We solve $(\theta^*, \omega^*,\tau^*) = \mathrm{argmin}_{\theta,\omega,\tau}\  Loss_{\theta,\omega,\tau}$, and the results of traffic state estimation and model parameter identification are presented in Table~\ref{tab:ch5-r3}. Empirically, the difference between true and estimated values is visually indistinguishable when the errors are smaller than 6.00 $\times 10^{-2}$ and 2.90 $\times 10^{-2}$ for density and velocity, respectively. Performances with accuracy below these values are considered as ``acceptable''. From Table~\ref{tab:ch5-r3}, we can observe that the TSE performance of PIDL+FDL with more than three loops is acceptable.

\begin{table}[h!]
    \caption{Performance of PIDL+FDL on the ARZ Dynamics}
    \label{tab:ch5-r3}
    \centering
    \begin{threeparttable}
    \begin{tabular}{ccccc}
    \toprule
    \toprule
    \multicolumn{1}{c}{$loops$} & \multicolumn{1}{c}{2} & \multicolumn{1}{c}{3} & \multicolumn{1}{c}{4} &
    \multicolumn{1}{c}{5}\\
    \midrule
    $Err(\hat{\rho},\rho)$    & 0.5111 & 0.2249 & 0.04871 & 0.05243\\ 
    $Err(\hat{u},u)$    & 0.1586 & 0.05914 & 0.01402 & 0.01389\\ 
    $\tau^*$    & 0.02276 & 0.018994 &  0.019654 & 0.021619\\
    \bottomrule
    \bottomrule
    \end{tabular}
    \begin{tablenotes}
      \footnotesize
      \item $loops$ stands for the number of loop detectors.  $\tau^*$ is the estimated relaxation time. Note the true $\tau=0.02$.
    \end{tablenotes}
  \end{threeparttable}
\end{table}


The experiment also demonstrates that the FDL performance improves as the number of loop detectors increases, and the proposed method with 4 loops and above is able to correctly learn the Greenshields FD relation. For visualization of the results, please refer to Figure~7 in the supplementary material in~\cite{Shi-arXiv-2021}.

\section{PIDL+FDL-Based TSE on NGSIM Data}
\label{sec-VI}

This section evaluates the PIDL+FDL-based TSE method using real-world traffic data, the  Next  Generation  SIMulation (NGSIM) dataset\footnote{www.fhwa.dot.gov/publications/research/operations/07030/index.cfm}, and compares its performance to baselines.

\subsection{NGSIM Dataset}

NGSIM dataset is widely-used and contains real-world vehicle trajectories on several road scenarios. We focus on a segment of the US Highway 101 (US 101), monitored by a camera mounted on top of a high building on June 15, 2005. The locations and actions of each vehicle in the monitored region for a total of around 680 meters and  2,770 seconds were converted from camera videos.

We select the data from all the mainline lanes of the US 101 highway segment to calculate the average traffic density for approximately every 30~meters over a 1.5~seconds period. After preprocessing to remove the time when there are non-monitored vehicles running on the road (at the beginning and end of the video), there are 21 and 1770 valid cells on the spatial and temporal dimensions, respectively. We treat the center of each cell as a grid point. 
The spatio-temporal field of traffic density $\rho(t,x)$ and velocity $u(t,x)$ in the dataset can be visualized in Figure 8 in the supplementary in~\cite{Shi-arXiv-2021}.


For TSE experiments in this section, loop detectors are used to provide observed data with a recording frequency of 1.5 seconds. By default, they are evenly installed on the highway segment.  We assume that the loop detectors are able to record the density and average velocity of cells on certain locations

\subsection{TSE Methods for Real Data}

We first introduce the PIDL+FDL-based methods for the real-world TSE problem, and then, describe advanced baselines to compare with.

\begin{figure}[h!]
\centering
  \includegraphics[scale=0.65]{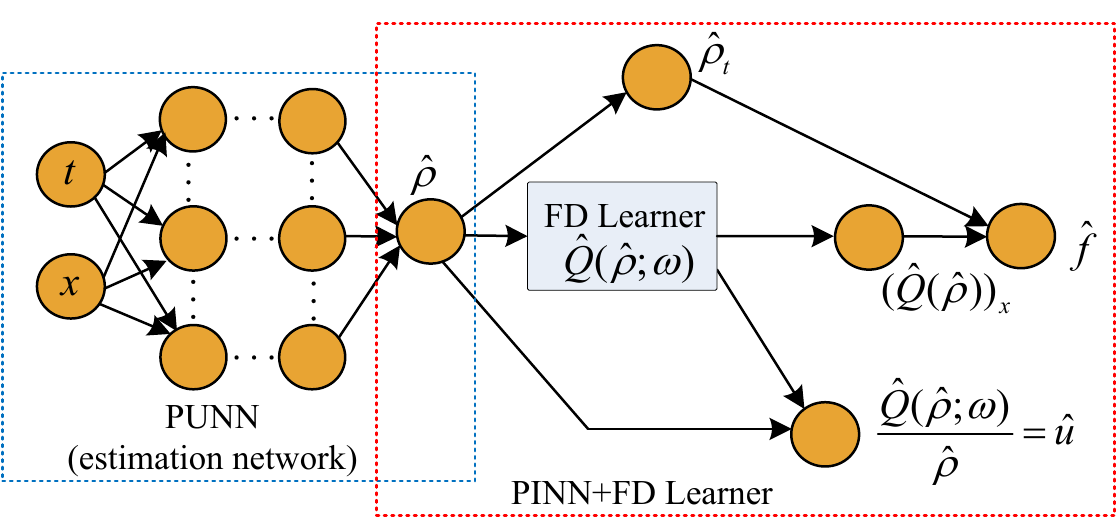}
  \caption{ The structure design of LWR-PIDL+FDL method for NGSIM data.}
  \label{fig:ch6-lwr-fdl}
\end{figure}

\textbf{LWR-PIDL+FDL}: This method is based on the PIDL+FDL encoded with the first-order LWR, using the structure in Fig.~\ref{fig:ch4-PINN_structure-r3} except for two modifications: 1) no assumptions on diffusion effects are made and the traffic flow becomes $\rho_t+(Q(\rho))_x=0$, which is commonly-used in literature~\cite{Fan-2013}; and 2) the estimation of velocity is calculated using an additional calculation node $\hat{u}=\hat{Q}(\hat{\rho};\omega)/\hat{\rho}$. Specifically, the modified structure of this TSE method is presented in Fig.~\ref{fig:ch6-lwr-fdl}. 
We select 80\% of the grid~$G$ as the auxiliary points $A$. The loss in Eq.~(\ref{equ-11-r3}) (with the MSE on velocity estimation added and the boundary conditions removed) is used for training the PUNN and FDL using the observed data from loop detectors (i.e., both observation points $O$ and corresponding target state values $P$). After tuning the hyperparameters with grid search, we present the minimal-achievable estimation errors. The same for other baselines by default. Because the real data could be noisy, leading to abnormal learned FD curves, the reshaping regularization term in Eq.~(\ref{equ-3-8-r3}) is applied.

\textbf{ARZ-PIDL+FDL}: This method is based on the PIDL+FDL encoded with the second-order ARZ, using the structure in Fig.~\ref{fig:ch4-ARZ-PIDL-r3}. The Eq.~(\ref{equ-15-r3}) (with boundary conditions removed) is applied as the loss function. Other experimental setups are the same with those of the LWR-PIDL+FDL method.

\textbf{Two-Dimensional Data Interpolation  (Interp2)}: The two-dimensional linear interpolation method is used as a baseline, which interpolates the traffic states using the neighboring observed data in a linear manner.

\textbf{Adaptive Smoothing (AS) Method}: This method estimates the traffic state of a cell using the sum of all the observed data weighted by some smoothing kernel filters. We implement a generalized AS method proposed in~\cite{Treiber-2011} with parameters suggested in~\cite{XudongWang-2021}.

\textbf{Long Short Term Memory (LSTM) based Method}: This baseline method employs the LSTM architecture, which is customized from the LSTM-based TSE proposed by~\cite{LiWei-2018}. This model can be applied to our problem by leveraging the spatial dependency, i.e., to use the information of previous cells to estimate the traffic density and velocity of the next cell along the spatial dimension. 

Other baselines include the \textbf{Pure Neural Network (NN)} and the \textbf{Extended Kalman Filter (EKF)} as well as the advanced PIDL-based TSE methods: \textbf{LWR-based PIDL (LWR-PIDL)} and \textbf{ARZ-based PIDL (ARZ-PIDL)}. For more descriptions regarding these four baselines, we refer the readers to the authors' previous work in~\cite{Shi-AAAI-2021}.

\subsection{Results and Discussion}

\begin{figure}[h!]
\centering
  \includegraphics[scale=0.72]{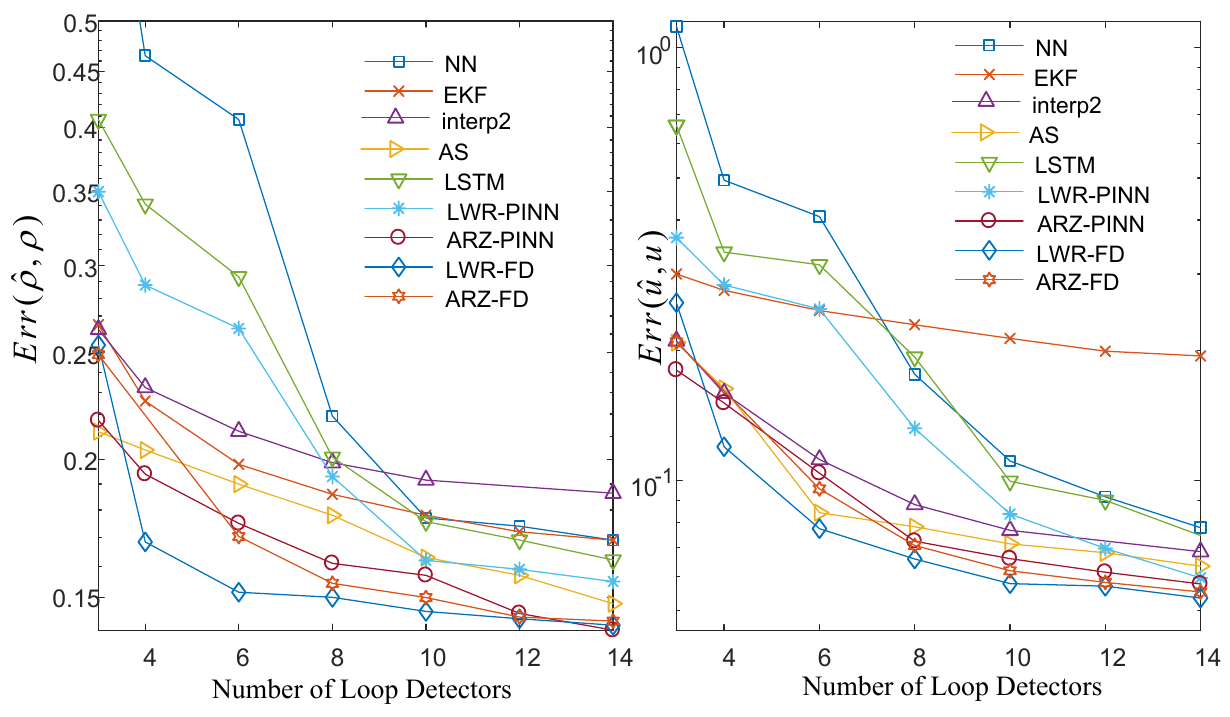}
  \caption{TSE performance of PIDL+FDL-based methods and baselines.}
  \label{fig:ch6-results-r3}
\end{figure}

We apply PIDL+FDL-based, and baseline methods to TSE on the NGSIM dataset with different numbers of loop detectors. The results are presented in Fig.~\ref{fig:ch6-results-r3}.

From Fig.~\ref{fig:ch6-results-r3}, we can observe that the PIDL-related methods generally perform better than the model-driven and data-driven TSE baselines. The EKF/Interp2/AS methods achieve better errors than the NN/LSTM methods when the number of loop detectors are small, while the NN/LSTM methods' performance catches up when more loops are available.

The results are reasonable. The EKF is a model-driven approach, making sufficient use of the traffic flow model to appropriately estimate unobserved values when limited data are available. However, the model cannot fully capture the complicated traffic dynamics in the real world, and as a result, the EKF's performance flattens out. The Interp2 is a non-parametric data-driven method interpolating the unobserved fields using neighboring observation in a linear manner. The AS method incorporates the characteristic velocities of information propagation in free and  congested  traffic, by  skewing  the  principal  axes  of  the smoothing kernel. The Interp2 and AS methods have a relatively low complexity which can prevent over-fitting when the data is small. However, they may not effectively handle subtle state changes in the unobserved area due to the linearly-extrapolating nature or filtering nature, respectively. The PIDL-based method's errors are generally below the baselines, because it can make efficient use of both the traffic flow model and observed data. The ARZ-PIDL method is informed by a more advanced second-order traffic model and its performance is superior to that of LWR-PIDL.

The PIDL+FDL-based methods can generally achieve the best estimation accuracy and data efficiency over the above TSE baselines. The results demonstrate that the proper integration of the NN-based FD surrogate to the PIDL can give the learning framework more flexibility to achieve an improved TSE accuracy. One interesting phenomenon is that the PIDL+FDL with the first-order LWR (LWR-PIDL+FDL) can beat the one with a more sophisticated second-order ARZ model (ARZ-PIDL+FDL). This observation supports our discussion that sophisticated traffic models may not always lead to a better TSE performance, because the model may contain complicated terms that makes the TSE performance sensitive to the PINN structural design, and thus, the model becomes difficult to train. Compared to the ARZ-PIDL+FDL, the LWR-PIDL+FDL can balance the trade-off between the sophisticated level of PINN and the training flexibility more properly, making it a better fit to the NGSIM scenario.

\subsection{Discussions on Fundamental Diagram Estimation}

\begin{figure}[h!]
\centering
  \includegraphics[scale=0.55]{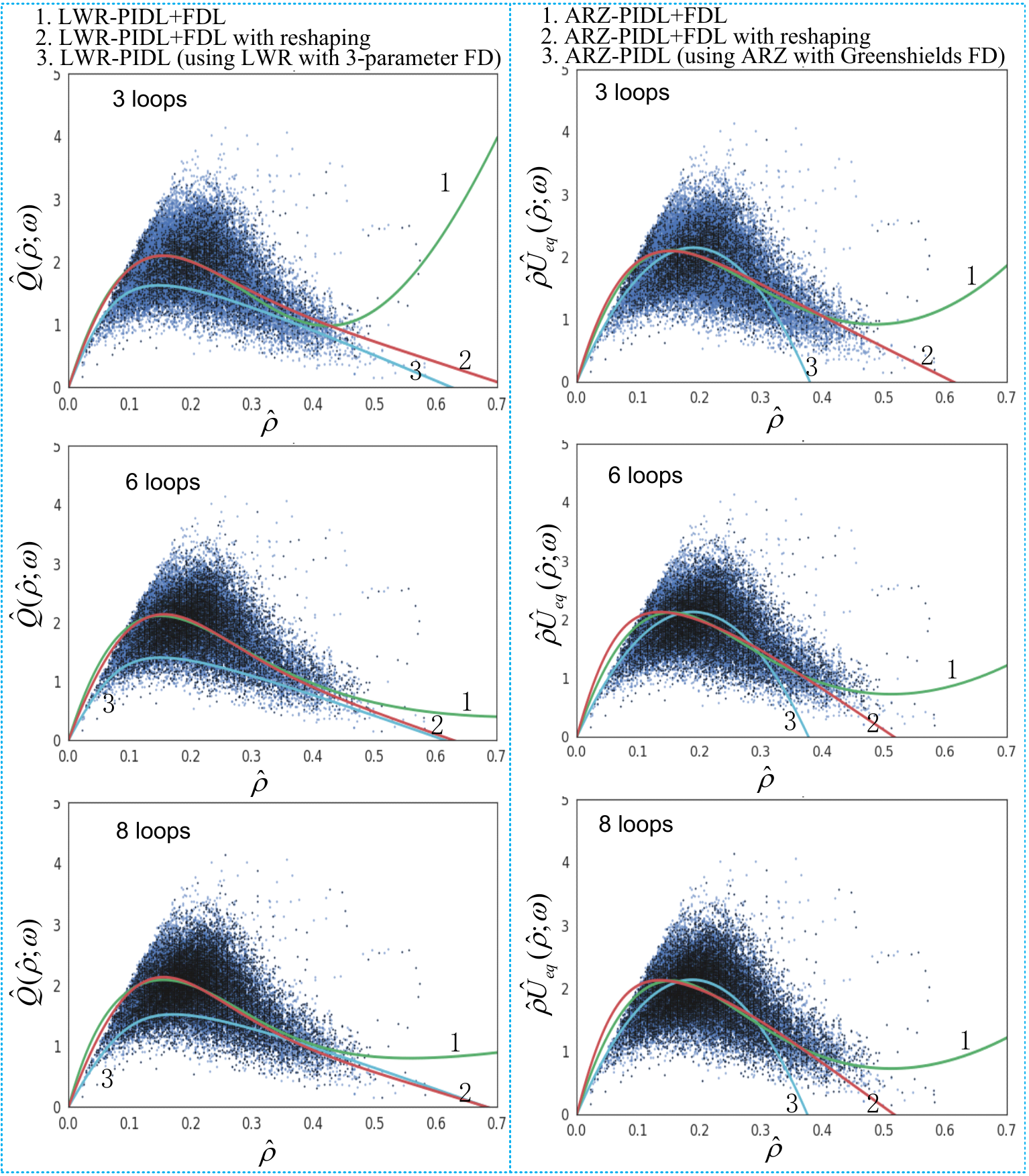}
  \caption{Comparison of the FD curves estimated by the PIDL+FDL-based TSE methods and PIDL-based TSE methods. For consistent visualization, the learned $\hat{U}_{eq}(\hat{\rho};\omega)$ is converted to $\hat{\rho}\hat{U}_{eq}(\hat{\rho};\omega)$ to represent the estimated flow.}
  \label{fig:ch6-FD}
\end{figure}

The PIDL+FDL-based methods can further learn the hidden fundamental diagram (FD) relation. We compare the FD curves learned via the PIDL+FDL-based methods and PIDL-based methods in the density-flow space when small number of loops are available. The results are presented in Fig.~\ref{fig:ch6-FD} where each dark blue dot is a density-flow data point in the NGSIM dataset. Note, the flow values are not part of the observed data during the training phase.

For the PIDL-based methods, the closed form of the flux function and velocity function are given and the parameters in the PINN component are learned along with the TSE training, and thus, the shape of the FD curves are predefined. The corresponding FD curves with the learned model parameters are indexed as  ``3'' in Fig.~\ref{fig:ch6-FD}. For consistent visualization, the learned $\hat{U}_{eq}(\hat{\rho};\omega)$ is converted to $\hat{\rho}\hat{U}_{eq}(\hat{\rho};\omega)$ to represent the estimated flow. The LWR-PIDL method is encoded with the 3-parameter-based flux, and the ARZ-PIDL is with the Greenshields function for the equilibrium velocity. The former has a proper shape defined by the given mathematical formula, but due to the complicated nature of the PINN for the 3-parameter flux and the noisy quality of the data, the learned FD curves do not fit the density-flow points to a satisfactory extent. The latter has a predefined quadratic shape and can capture the density-flow characteristics to a limited level.

For the PIDL+FDL-based methods, using the standard training loss in Eq.(\ref{equ-3-7-r3}),  the learned FD shapes (indexed as ``1'' in Fig.~\ref{fig:ch6-FD}) fit the NGSIM density-flow points well over the domain where the observed traffic state data are distributed, i.e., around $\rho \in [0,0.5]$. However, the FD curves tend to curl up in the large-density domain, where the data are sparse. To address this abnormal shape, we apply the regularization term in Eq.~(\ref{equ-3-8-r3}) for reshaping and impose the prior knowledge of concavity over a narrowed interval of $\rho \in [0.6,0.7]$. To this end, we set the hyperparameters in Eq.~(\ref{equ-3-8-r3}) to $a=0.6$ and $b=0.7$. The corresponding learned FD curves using the reshaping regularization term are indexed as ``2'' in the figures, and they can properly capture the density-flow characteristics to a satisfactory level. Because of using the FD Learner, the LWR-PIDL+FDL contains no model parameters, and the conservation law plus the $\hat{Q}(\rho;\omega^*)$ constitutes the LWR model reconstructed by the LWR-PIDL+FDL. The ARZ-PIDL+FDL contains one model parameter, i.e., the relaxation time $\tau$. The learning with data from 3, 6 and 8 loops converges to $\tau^*=[23.36,25.99,27.66]$, which is reasonably close to $\tau=[27.6,28.8,30.5]$ directly fitted from data. The conservation law, the momentum of velocity with model parameter $\tau^*$, and the learned $\hat{U}_{eq}(\rho;\omega^*)$ constitute the ARZ model reconstructed by the ARZ-PIDL+FDL. 

The results demonstrate that the proposed PIDL+FDL-based TSE method (with the regularization for reshaping) is able to efficiently conduct high-quality TSE, model parameter identification and fundamental diagram estimation at the same time with relatively small amounts of observed data.

\section{Conclusion}

We introduced the PIDL+FDL framework to the TSE problem on highways using loop detector data and demonstrate the significant benefits of the integration of an ML surrogate into the model-driven component in PIDL. This framework can be used to handle traffic state estimation, model parameter identification, and fundamental diagram estimation simultaneously. The experiments on real highway data show that PIDL+FDL-based approaches can outperform baselines in terms of estimation accuracy and data efficiency as well as the estimation of FD.

The limitations and potential future works of this paper are as follows: (1) Similar to most deep learning methods, hyperparameter tuning is an issue of PIDL-based TSE, and tuning such a large number of hyperparameters based on approaches like cross-validation is too complicated for real-world application, and the model basically has to be tuned for each scenario (spatial resolution, temporal resolution, spatial grid size, temporal grid size, observation error, etc), which limits its applicability to real-world problems;
(2) PIDL is known to have issues with noisy data, and how the high noise and corruption in the real traffic data undermine the performance of PIDL-based TSE needs further investigations;
(3) It is worthy of considering more ML surrogate components to represent other unobserved traffic quantities in the traffic flow model, such as $h(\rho)$ and $U_{eq}(\rho-u)/\tau$, 
and study to what extent the addition of surrogates affects the performance. 

\section*{Acknowledgment}

This work is partially supported by NSF DMS-1937254, DMS-2012562 and CCF-1704833. 
We would like to thank Professor Benjamin Seibold from Temple University for inspiring us on FDL.

\ifCLASSOPTIONcaptionsoff
  \newpage
\fi



%



\bibliographystyle{IEEEtran}

\bibliography{IEEEabrv, references.bib}

\vspace{5mm}

%

\end{document}